\documentclass[conference]{IEEEtran}
\IEEEoverridecommandlockouts
\usepackage{cite}
\usepackage{amsmath,amssymb,amsfonts}
\usepackage{algorithmic}
\usepackage{graphicx}
\usepackage{textcomp}
\usepackage{xcolor}
\usepackage{balance}
\usepackage{hyperref}
\usepackage[symbol]{footmisc}
\usepackage[]{collab}

\collabAuthor{zb}{teal}{Zhuangbin Chen}
\collabAuthor{yx}{blue}{Yuxin Su}
\collabAuthor{hy}{red}{HY}

\def\BibTeX{{\rm B\kern-.05em{\sc i\kern-.025em b}\kern-.08em
    T\kern-.1667em\lower.7ex\hbox{E}\kern-.125emX}}
\begin{document}

\title{Graph-based Incident Aggregation for Large-Scale Online Service Systems}
% \title{Conference Paper Title*\\

% \author{Anonymous author(s)}

\author{\IEEEauthorblockN{Zhuangbin Chen$^{\ast}$, Jinyang Liu$^{\ast}$, Yuxin Su$^{\ast\dag}$\thanks{$^{\dag}$Corresponding author.}, Hongyu Zhang$^{\ddag}$ \\Xuemin Wen$^{\P}$, Xiao Ling$^{\P}$, Yongqiang Yang$^{\P}$, Michael R. Lyu$^{\ast}$}
\IEEEauthorblockA{$^{\ast}$The Chinese University of Hong Kong, Hong Kong, China, \{zbchen, jyliu, yxsu, lyu\}@cse.cuhk.edu.hk \\
$^{\ddag}$The University of Newcastle, NSW, Australia, hongyu.zhang@newcastle.edu.au \\
$^{\P}$Huawei, China, \{wenxuemin, lingxiao1, yangyongqiang\}@huawei.com}

% \and
% \IEEEauthorblockN{Jinyang Liu, Yuxin Su\thanks{Yuxin Su is the corresponding author.}}
% \IEEEauthorblockA{\textit{The Chinese University of Hong Kong} \\
% Hong Kong, China \\
% \{jyliu,yxsu\}@cse.cuhk.edu.hk}
% \and
% \IEEEauthorblockN{Hongyu Zhang}
% \IEEEauthorblockA{\textit{The University of Newcastle} \\
% NSW, Australia \\
% hongyu.zhang@newcastle.edu.au}
% \and
% \IEEEauthorblockN{Xuemin Wen, Xiao Ling, Yongqiang Yang}
% \IEEEauthorblockA{\textit{Computing and Networking Innovation Lab, Cloud BU, Huawei} \\
% China \\
% \{wenxuemin,lingxiao1,yangyongqiang\}@huawei.com}
% \and
% \IEEEauthorblockN{Yongqiang Yang}
% \IEEEauthorblockA{\textit{Huawei} \\
% China \\
% yangyongqiang@huawei.com}
% \and
% \IEEEauthorblockN{Michael R. Lyu}
% \IEEEauthorblockA{\textit{The Chinese University of Hong Kong} \\
% Hong Kong, China \\
% lyu@cse.cuhk.edu.hk}
}

% \and
% \and
% \and
% \and
% \and

\maketitle

\begin{abstract}

As online service systems continue to grow in terms of complexity and volume, how service incidents are managed will significantly impact company revenue and user trust. Due to the cascading effect, cloud failures often come with an overwhelming number of incidents from dependent services and devices. To pursue efficient incident management, related incidents should be quickly aggregated to narrow down the problem scope. To this end, in this paper, we propose GRLIA, an incident aggregation framework based on graph representation learning over the cascading graph of cloud failures. A representation vector is learned for each unique type of incident in an unsupervised and unified manner, which is able to simultaneously encode the topological and temporal correlations among incidents. Thus, it can be easily employed for online incident aggregation. In particular, to learn the correlations more accurately, we try to recover the complete scope of failures' cascading impact by leveraging fine-grained system monitoring data, i.e., Key Performance Indicators (KPIs). The proposed framework is evaluated with real-world incident data collected from a large-scale online service system of Huawei Cloud. The experimental results demonstrate that GRLIA is effective and outperforms existing methods. Furthermore, our framework has been successfully deployed in industrial practice.

\end{abstract}

\begin{IEEEkeywords}
Cloud computing, online service systems, incident management, graph representation learning
\end{IEEEkeywords}

\section{Introduction}

In recent years, IT enterprises started to deploy their applications as online services on cloud, such as Microsoft Azure, Amazon Web Services, and Google Cloud Platform. These cloud computing platforms have benefited many enterprises by taking over the maintenance of IT and infrastructure and allowing them to improve their core business competence. However, for large-scale online service systems,
% \hy{here, it is cloud platforms or cloud applications? online service systems? Microservice systems?}
failures are inevitable, which may lead to performance degradation or service unavailability. Whether or not the service failures are properly managed will have a great impact on the company's revenue and users' trust. For example, an hour episode of downtime in Amazon led to a loss of 100+ million dollars~\cite{downtimeexample}.

% \hy{why abstractions? what is the definition of incident in this paper. The differences between failure and incident should be made clearly - some people may treat them the same. I think this is important for this paper},

% \textcolor{red}{In this paper, we define $incidents$ as system issues associated with service $failures$.}
When a failure happens, system monitors will render a large number of incidents to capture different failure symptoms~\cite{chen2020towards,chen2020identifying,zhao2020understanding}, which can help engineers quickly obtain a big picture of the failure and pinpoint the root cause. For example, %``OS filesystem readonly'' is a critical failure in the ELB (Elastic Load Balance) service and the incident
``Special instance cannot be migrated" is a critical network failure in Virtual Private Cloud (VPC) service, and the incident ``Tunnel bearing network pack loss'' is a signal for this network failure, which is caused by the breakdown of a physical network card on the tunnel path. %\hy{better relate the incident to the failure: do you have the name of that network failure? or do you have an incident for the ``OS filesystem readonly'' failure?}\hy{can you hep fill in the xxx}\yx{done}
% \hy{any examples of incidents for a failure? like the ones in Table 1}\yx{modified}
Due to the large scale and complexity of online service systems, the number of incidents is overwhelming the existing incident management systems~\cite{chenaiops,chen2020towards,zhao2020understanding}.
% \hy{can give a rough incident number/scale here}
% zb: I cannot find a reference, companies are not willing to release the numbers
When a service failure occurs, aggregating related incidents can greatly reduce the number of incidents that need to be investigated. For example, linking incidents that are caused by a hardware issue can provide engineers with a clear %profile 
picture of the failure, e.g., the type of the hardware error or even the specific malfunctioning components. Without automated incident aggregation, engineers may need to go through each incident to %realize
discover the existence of such a problem and collect all related incidents to understand it. Moreover, incident aggregation can also facilitate failure diagnosis. In cloud systems, some trivial incidents are being generated continuously, and multiple (independent) failures can happen at the same time. Identifying correlated incidents can therefore accelerate the process of root cause localization.

% \hy{say something like "To identify correlated incidents, we could measure the text similarity between two incident reports".}
% \yx{introduce common approach with textual similarity first}

To aggregate related incidents, one straightforward way is to measure the text similarity between two incident reports~\cite{zhao2020understanding,chen2020identifying}. For example, incidents that share similar titles are likely to be related. Besides textual similarity, system topology (e.g., service dependency, network IP routing) is also an important feature to resort to. Due to the dependencies among online services, failures often have a cascading effect on other inter-dependent services. A service dependency graph can help track related incidents caused by such an effect.
% However, the impact of a failure may not manifest itself completely over the system topology.
However, as cloud systems often possess certain ability of fault tolerance, some services may not report incidents, impeding the tracking of failures' impact (to be explained in Section~\ref{sec:background_problem_statement}).
% One important reason is that monitors \hy{briefly explain monitors, such as monitors} (the tools that continuously check the status of the cloud system) that report incidents are configured with predefined rules by engineers\hy{maybe this can go to background section}. The impact of the failure may not be strong enough to hit the threshold. Recently studies on incident management~\cite{chen2020towards,huang2017gray} has demonstrated the incompleteness and imperfection of monitor design and distribution in online service systems. Therefore, along the service dependency chain, some services in the middle may remain silent, which impedes the tracking of failure's cascading effect. 
This issue is ubiquitous in production systems, which has not yet been properly addressed in existing work. Moreover, the patterns of incidents are collectively influenced by different factors, such as their topological and temporal locality. Existing work~\cite{zhao2020understanding,chen2020identifying} combine them by a simple weighted sum, which may not be able to reveal the latent correlations among incidents.

%To address this problem, 
In this work, we propose GRLIA (stands for Graph Representation Learning-based Incident Aggregation), which is an incident aggregation framework to assist engineers in failure understanding and diagnosis. %Considering that cloud resources are structured in a graph form, We leverage  to learn a representation for each type of incident.
Different from the existing work of alert storm handling~\cite{zhao2020understanding} and linked incident identification~\cite{chen2020identifying}, we do not rely on incidents' textual similarity. Moreover, we learn incidents' topological and temporal correlations in a unified manner (instead of by a weighted combination). Traditional applications of graph representation learning often learn the semantics of a fixed graph. Unlike them, we propose to learn a feature representation for each unique type of incident,
% \hy{can revise: the object in our framework that carries a representation is incident},
which can appear in multiple places of the graph. The representation encodes the historical co-occurrence of incidents and their topological structure. Thus, they can be naturally used for incident aggregation in online scenarios. To track the impact graph of a failure (i.e., the incidents triggered by the failure), we exploit more fine-grained system signals, i.e., KPIs, as a piece of auxiliary information to discover the scope of its cascading effect. KPIs profile the impact of failures in a more sophisticated way. Therefore, if two services exhibit similar abnormal behaviors (characterized by incidents and KPIs), they should be suffering from the same problems even if no incidents have been reported. Finally, we apply community detection algorithms to find the scope of different failures.

%\hy{say some about experiments and results}
% Particularly, we cast the calculation of time series similarity into detecting their Granger causality~\cite{granger1969investigating}, which is able to deal with the problem of temporal drift and noise of KPI data.

To sum up, this work makes the following major contributions:

\begin{itemize}
    \item We propose to identify service failures' impact graph, which consists of the incidents that originate from the same failures. Such an impact graph helps us obtain a complete picture of failures' cascading effect.
    % To measure the consistency of system components' abnormal behaviors\hy{what is "the consistency of .. abnormal behaviors" and why to measure it, just say "measure system components' abnormal behaviors"?},
    To this end, we combine incidents with KPIs to measure the behavioral similarity between services. Community detection algorithms are then applied %to such similarity graph (to be discussed in detail in Section~\ref{sec:failure_impact_graph_identification})
    % \hy{similarity graph is not descried in the paper}
    %\hy{what is similarity graph? do you mean 'failure-impact graph'?}
    % \hy{no 'KPI similarity graph' is mentioned in the rest of the paper?}
    to determine the failure-impact graph of different failures automatically.
    
    \item We propose GRLIA, an incident aggregation framework based on graph representation learning. The embedding vector for each unique type of incident is learned in an unsupervised and unified fashion, which encodes its interactions with other incidents in temporal and topological dimensions. Online incident aggregation can then be naturally performed by calculating their distance. The implementation of GRLIA is available on GitHub~\cite{grlia}.
    
    \item We conduct experiments with real-world incidents collected from Huawei Cloud, which is a large-scale cloud service provider. The results demonstrate the effectiveness of the proposed framework. Furthermore, our framework has been successfully incorporated into the incident management system of Huawei Cloud. Feedback from on-site engineers confirms its practical usefulness. 
\end{itemize}

The remainder of this paper is organized as follows. Section~\ref{sec:background_problem_statement} introduces the background and problem statement of this paper. Section~\ref{sec:methodology} describes the proposed framework. Section~\ref{sec:evaluation} shows the experiments and experimental results. Section~\ref{sec:discussion} presents our success story and lessons learned from practice. Section~\ref{sec:related_work} discusses the related work. Finally, Section~\ref{sec:conclusion} concludes this work.
\section{Background and Problem Statement}
\label{sec:background_problem_statement}

\subsection{Topology of Large-scale Online Service Systems}

%Cloud enterprises
Cloud vendors provide a variety of online services to customers worldwide. In general, there are three main models of cloud-based services, %in cloud computing, 
namely, Software as a Service (SaaS), Platform as a Service (PaaS), and Infrastructure as a Service (IaaS)~\cite{kavis2014architecting}. 
% Large-scale online service systems \hy{the topology should be common to all cloud-based service systems, not just the system that we consider in this paper? the problem and the approach should be general...} are often constructed by following the service-oriented architecture (SOA)~\cite{erl1900service}, which designs infrastructure from the perspective of services instead of servers.
%In a large-scale online service system, services \hy{not clear: Services in an SOA. not sure if the concept of SOA is really needed here. can just say "The cloud-based service systems have a hierarchical topology?} \zb{I removed SOA} can serve as one of the above three service models. Therefore, 
Online service systems often possess a hierarchical topology, i.e., the stack of application, platform, and infrastructure layers. Each service embodies the integration of code and data required to execute a complete and discrete functionality. For example, in the application layer, the services provided to customers can be a user application, a microservice, or even a function; in the platform layer, the services can be a container or a database; in the infrastructure layer, the services can be a virtual machine or storage. Different services communicate through virtual networks using protocols such as Hypertext Transfer Protocol (HTTP) and Remote Procedure Call (RPC). Such communications among services constitute the complex topology of large-scale online service systems.

\begin{table*}
    \centering
    \caption{Examples of incident aggregation
    % Incidents in blue are related to a network failure; incidents in gray are caused by a hardware problem (disk error).
    }
    \includegraphics[width=0.86\linewidth]{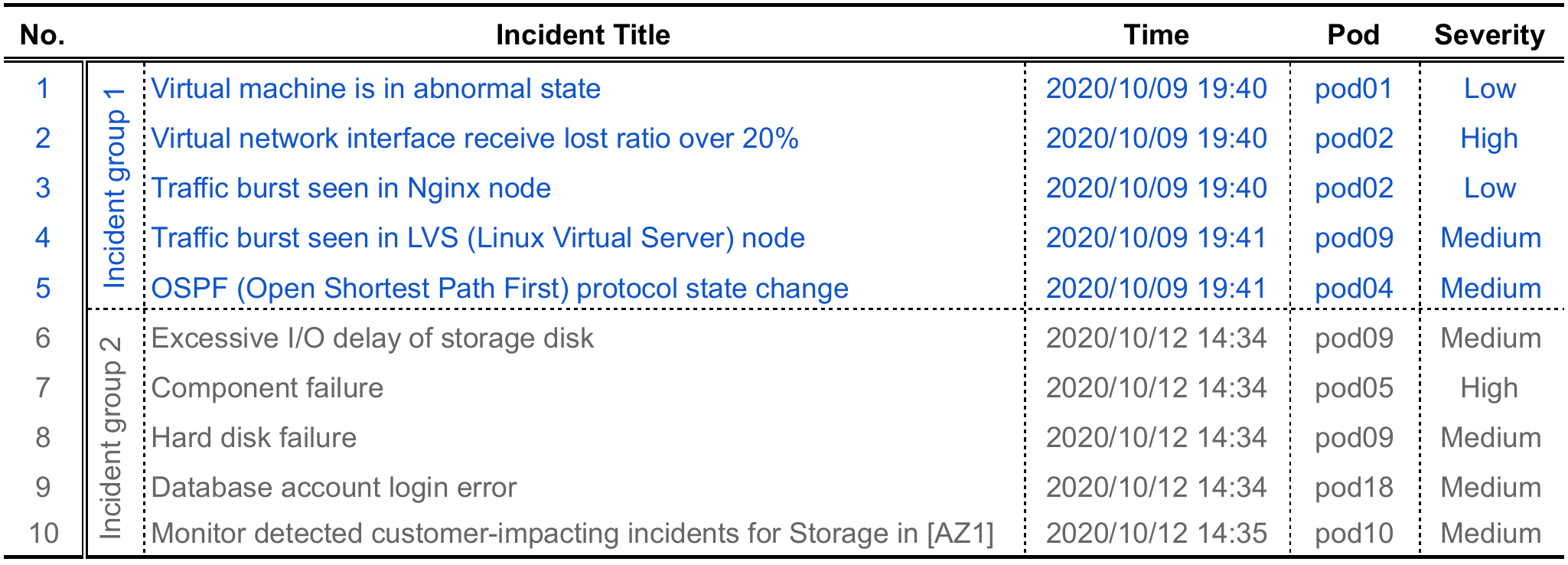}
    \label{fig:incident_aggregation_example}
\end{table*}

\begin{figure}
    \centering
    \includegraphics[width=1.0\linewidth]{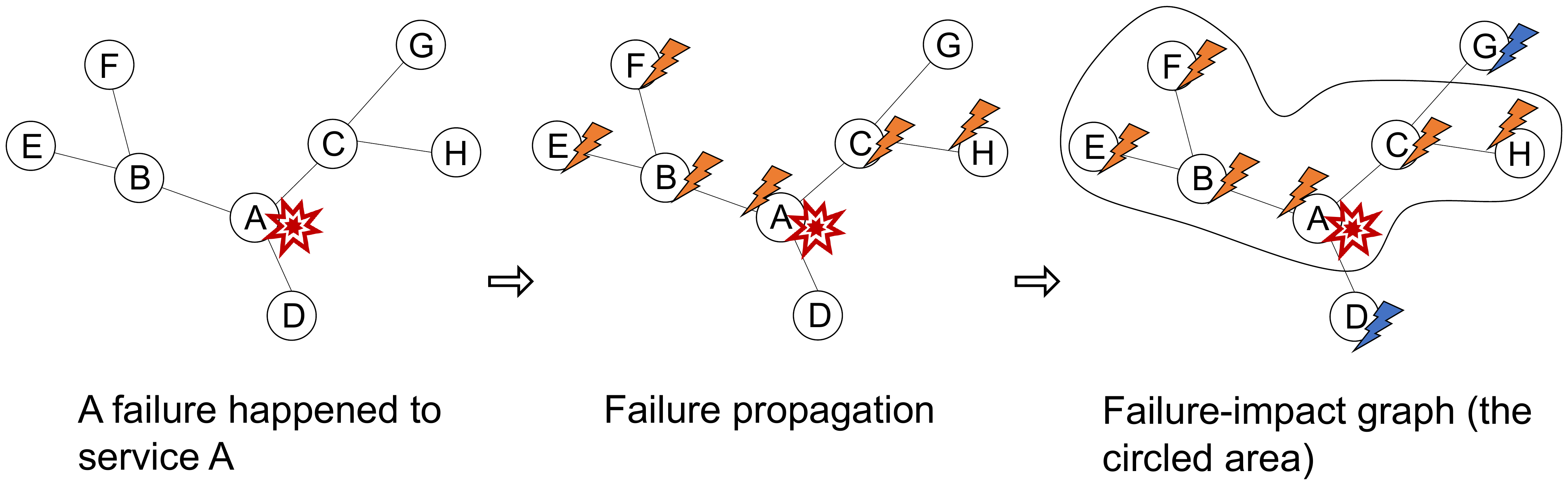}
    \caption{An illustration of service failures' cascading effect. The irregular circle in the third subfigure shows the failure-impact graph.}
    \label{fig:cascading_topo}
\end{figure}

\begin{figure}
    \centering
    \includegraphics[width=0.9\linewidth]{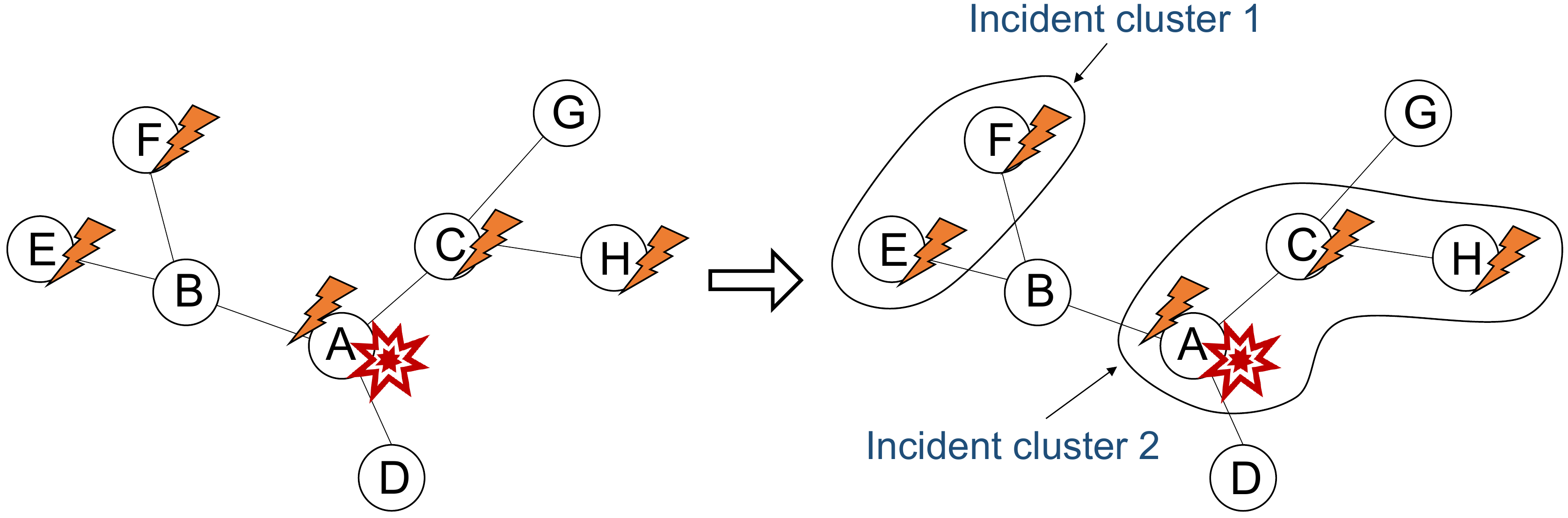}
    \caption{An example of incomplete failure-impact graph}
    \label{fig:incomplete_impact_graph}
\end{figure}

\subsection{Cascading Effect of Service Failures}
\label{sec:cascading_effect}

% \hy{can describe more about monitors, incidents, and failures here.}
With such a topology, a failure occurring to one service tends to have a cascading effect across the entire system. Representative service failures include slow response, request timeout, service unavailability, etc., which could be caused by capacity issues, configuration errors, software bugs, hardware faults, etc. To quickly understand failure symptoms, a large number of monitors are configured to monitor the states of different services in a cloud system~\cite{chen2020towards}. A monitor will render an incident when certain predefined conditions (e.g., ``CPU utilization rate exceeds 80\%'') are met. Typical configurations of monitors include setting thresholds for specific metrics (e.g., RPC latency, error counter), checking service/device availability or status, etc. When a failure happens, the monitors often render a large number of incidents. These incidents are triggered by the common root cause and describe the failure from different aspects. Thus, they can be aggregated to help engineers understand and diagnose the failure.
%, which describes the issues that the services are experiencing. For example, %``Intermittent availability of SQL service reported by users'', ``CPU utilization rate exceeds 80\%''.
% the health services along the dependency chain may also report incidents.

In this paper, we model the set of incidents triggered by a failure as its impact graph (or failure-impact graph), as illustrated by Fig.~\ref{fig:cascading_topo}.
%, i.e., they share a common root cause.
%Fig.~\ref{fig:cascading_topo} illustrates such concept.} 
Specifically, service $A$ encounters a failure, and the impact propagates to other services along the system topology. The circled area indicates the impact graph of the failure, where irrelevant incidents (in blue) in service $D$ and $G$ are excluded. In general, the system topology can have many different forms, such as the dependencies of services~\cite{ma2018using}, the configured IP routing of a cloud network~\cite{natarajan2012nsdminer}, etc. Intuitively, it might seem that the impact graph can be easily constructed by tracing incidents along the system topology. However, our industrial practices reveal that they are usually incomplete. An example is given in Fig.~\ref{fig:incomplete_impact_graph}, where service $B$ occasionally fails to report any incident during the failure. Existing approaches may perceive it as two separate failures, which is undesirable. We have summarized the following two main reasons for the missing incidents:

\begin{itemize}
    \item System monitors that report incidents are configured with rules predefined by engineers. Due to the diversity of cloud services and user behaviors, the impact of a failure may not meet the rules of some monitors. For example, if an application triggers an incident when its CPU usage rate exceeds 80\%, then any value below the threshold will be unqualified. As a consequence, the monitors will not report any incident, and thus, the tracking of the failure's impact is blocked. %Nevertheless, it is not saying the failure poses no impact on the server. For example, spikes are likely to be observed in its CPU usage curve (i.e., anomalies).
    
    \item To ensure the continuity of online services, %mission-critical applications,
    cloud systems are designed to have a certain fault tolerance capability. In this case, service systems can bear some abnormal conditions, and thus, no incidents will be reported. Therefore, the impact of a failure may not manifest itself completely over the system topology.
\end{itemize}

% \hy{maybe this can go to background section, need merging:}
% One important reason is that monitors (the tools that continuously check the status of the cloud system) that report incidents are configured with predefined rules by engineers.
% The impact of the failure may not be strong enough to hit the threshold.
Recent studies on cloud incident management~\cite{chen2020towards,huang2017gray} have demonstrated the incompleteness and imperfection of monitor design and distribution in online service systems. Thus, along the service dependency chain, some services in the middle may remain silent (i.e., report no incident), which impedes the tracking of failure’s cascading effect. Therefore, although online service systems generate many incidents, they are often scattered.  
% Correlating incidents can help us recover the big picture of service failures thus accelerate failure diagnosis.

\begin{figure*}
    \centering
    \includegraphics[width=1.0\linewidth]{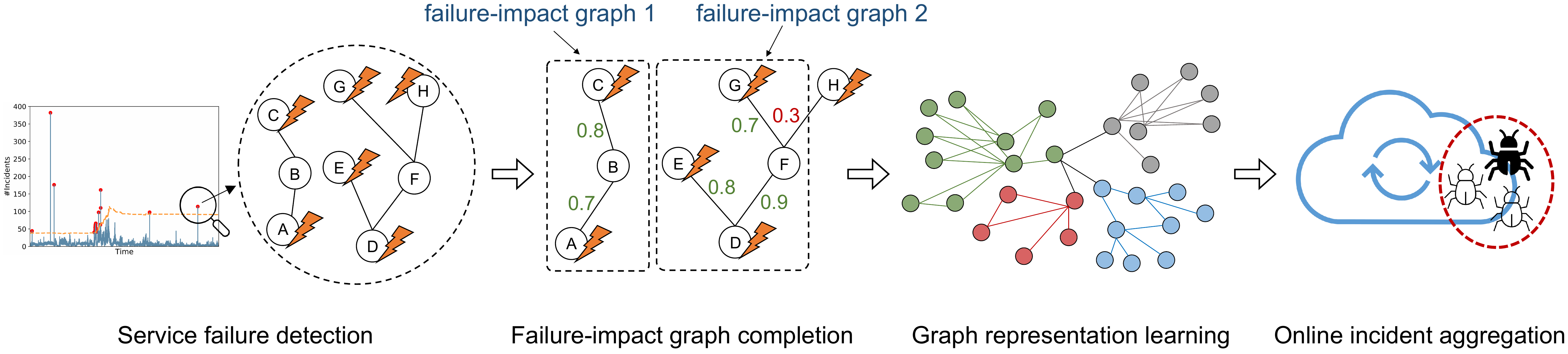}
    \caption{The overall framework of GRLIA}
    \label{fig:incident_graph_framework}
\end{figure*}

\subsection{Problem Statement}

This work aims to assist engineers in failure understanding and diagnosis with online incident aggregation, which is to aggregate incidents caused by the common failure. When services encounter failures, incidents that capture different failure symptoms constitute an essential source for engineers to conduct a diagnosis. However, it is time-consuming and tedious for engineers to manually examine each incident for failure investigation when faced with such an overwhelming number of incidents. Online incident aggregation is to cluster relevant incidents when they come in a streaming manner (i.e., continuously reported by the system). Examples are presented in Table~\ref{fig:incident_aggregation_example}, where items in blue and gray belong to two groups of aggregated incidents. Particularly, the first group shows a virtual network failure. Note that only the No.3 and No.4 incidents share some words in common, while the others do not. Meanwhile, the second group describes a hardware failure, and more specifically, a storage disk error. Engineers can benefit from such incident aggregation as the problem scope is narrowed down to each incident cluster.
%\hy{can briefly say why the first 4 incidents are in the same group, and the rest are in the other group. What are the root cause, etc.}

However, accurately aggregating incidents %\hy{here the word "online" is missing, make it clear if it is Online incident aggregation or Offline incident aggregation, or both. In previous definition, it was Online incident aggregation} 
for online service systems is challenging. We have identified three main reasons.

\textbf{Background noise}. Although related incidents are indeed generated around
%at, or close-to, 
the same time, many other cloud components are also constantly rendering incidents. These incidents are mostly trivial issues and therefore become background noise. Incident aggregation based on temporal similarity would suffer from a high rate of false positives.

\textbf{Dissimilar textual description}. Text (e.g., incident title and summary) similarity is an essential metric for incident correlation, which has been widely used in existing work~\cite{zhao2020understanding,chen2020identifying}. However, in reality, related incidents, especially the critical ones, do not necessarily have similar titles. Failing to correlate such critical incidents greatly hinders root cause diagnosis. %localization.

\textbf{Unclear failure-impact graph}. To correlate incidents accurately, we need to estimate the impact graph of service failures. As discussed in Section~\ref{sec:cascading_effect}, this task is challenging. Incidents alone are insufficient to completely reflect the impact of failures on the entire system. Therefore, we need to utilize more fine-grained information of the failures.

% Therefore, to achieve a better performance, we leverage different sources of monitoring data and the topology of cloud services.
\section{Methodology}
\label{sec:methodology}

\subsection{Overview}

% In online scenarios, incidents are constantly generated by cloud service systems. %or devices. 
% When cloud failures happen, especially critical ones, engineers will receive an overwhelming number of incidents. It would be labor-intensive and tedious to manually investigate each incident for failure diagnosis. GRLIA facilitates this process by providing automated incident aggregation.

In cloud systems, a large number of monitors are configured to continuously monitor the states of its services %and infrastructure 
from different aspects. Many incidents rendered by the monitors tend to co-occur due to their underlying dependencies. For example, some failure symptoms often appear together, and some incidents may develop causal relationship. Our main idea is to capture the co-occurrences among incidents by learning from historical failures. In online scenarios, such correlations can be leveraged to distinguish correlated incidents that are generated in streams.

The overall framework of GRLIA is illustrated in Fig.~\ref{fig:incident_graph_framework}, which consists of four phases, i.e., \textit{service failure detection}, \textit{failure-impact graph completion}, \textit{graph representation learning}, and \textit{online incident aggregation}. The first phase tries to identify the occurrence of service failures and retrieves different types of monitoring data, including incidents, KPI time series, and service system topology.
% The specific type of the topology can vary depending on the application scenario
In the second phase, we try to identify the incidents that are triggered by each individual failure detected above. More often than not, it is hard to precisely identify the impact scope of failures (as discussed in Section~\ref{sec:cascading_effect}), which hinders the learning of incidents' correlations. Therefore, we utilize the trends observed in KPI curves to auto-complete the failure-impact graphs. After obtaining the set of incidents associated with each failure, in the third phase, an embedding vector is learned for different types of incidents by leveraging existing graph representation learning models~\cite{hamilton2017representation,zhou2018graph}. Such representation encodes not only the temporal locality of incidents, but also their topological relationship. In the final phase, the learned incident representation will be employed for online incident aggregation by considering their cosine similarity and topological distance. In particular, we do not explicitly consider the dynamic change of a system topology because the changes often happen to a small area of the topology, e.g., container creation or kill. GRLIA essentially learns the correlations among incidents, which are also applicable to the changed portion of the topology. Nevertheless, when the system topology goes through a significant alteration, our framework is efficient enough to support quick model retraining.
% The source code is publicly available on Github\footnote{https://github.com/dsn21-85/code}.

\subsection{Service Failure Detection}

% \textcolor{red}{Change it to something like unhealthy system state detection, which are not necessarily cloud failures. However, they provide a good opportunity to study the interactions between incidents.} 
Due to the cascading effect, when service failures occur, a large number of incidents are often reported in a short period of time.
% However, when incidents are reported, it does not necessarily indicate the occurrence of a failure. In fact, most of the time, they are only alerting trivial system issues.
Thus, setting a fixed threshold for the average number of reported incidents (e.g., \#incidents/min$>$50) could be a reasonable criterion to detect failures. However, such a design suffers from a trade-off between false positives and false negatives due to online service systems' complex and ever-changing nature~\cite{zhao2020understanding}. For example, different services have distinct sensitivity to the number of incidents, and continuous system evolution/feature upgrades could change the threshold. Thus, a self-adaptive algorithm is more desirable.

For time-series data, anomalies often manifest themselves as having a large magnitude of upward/downward changes. Extreme Value Theory (EVT)~\cite{siffer2017anomaly} is a popular statistical tool to identify data points with extreme deviations from the median of a probability distribution. It has been applied to predicting unusual events, e.g., severe floods and tornado outbreaks~\cite{de2007extreme}, by finding the law of extreme values that usually reside at the tail of a distribution. Moreover, it requires no hand-set thresholds and makes no assumptions on data distribution. In this work, we follow~\cite{siffer2017anomaly,zhao2020understanding} to detect bursts %in steaming series of per-minute number of incidents, which
in time series of the number of incidents per minute. As a typical time series anomaly detection problem, other approaches (e.g.,~\cite{lin2018predicting,hundman2018detecting}) in this field are also applicable. The bursts are regarded as the occurrence of service failures. This algorithm can automatically learn the normality of the data in a dynamic environment and adapt the detection method accordingly. Fig.~\ref{fig:incident_graph_framework} (phase one) presents an example of service failure detection, where all abnormal spikes are successfully found by the decision boundary (the orange dashed line). For consecutive bins that are marked as anomalies, we regard them as one failure because failures may last for more than one minute. The next phase will distinguish multiple (independent) failures that happen simultaneously. Particularly, the detection algorithm is only required to have a high recall, and the precision is of less importance. It is because the goal of the follow-up two phases is to find the correlations between incidents. Such correlation rules will not be violated even incidents are not appearing together during actual cloud failures.

\subsection{Failure-Impact Graph Identification}
\label{sec:failure_impact_graph_identification}

% \zb{Can we do the whole thing (incident aggregation) only with the techniques introduced in this step? Reasons could be: it is not accurate (vulnerable to noise); not efficient; does not consider the relationship between incident shown in the history.}

In the first phase, the number of incidents %of the entire system 
per minute is calculated, and incident bursts are regarded as the occurrence of service failures. For each failure, the %\hy{some incident reports??}
incidents collected from the entire system are not necessarily related to it. This is because: 1) while some services are suffering from the failure, others may continuously report incidents (could be trivial and unrelated issues); and 2) multiple service failures could happen simultaneously. Therefore, we need to identify the set of incidents for each individual failure that is generated due to the cascading effect.

To this end, the concept of community detection is exploited. Community detection algorithms aim to group the vertices of a graph into distinct sets, or communities, such that there exist dense connections within a community and sparse connections between communities. Each community represents a collection of incidents rendered by the common service failure, in which the correlations among incidents can be explored. A comparative review of different community detection algorithms is available in~\cite{yang2016comparative}. In this work, we employ the well-known \textit{Louvain} algorithm~\cite{blondel2008fast}, which is based upon modularity maximization. The modularity of a graph partition measures the density of links inside communities compared to links between communities. For weighted graphs, the modularity can be calculated as follows~\cite{blondel2008fast}:

\begin{equation}
    M=\frac{1}{2m}\sum_{i,j}[W_{i,j}-\frac{k_ik_j}{2m}]\delta(c_i, c_j)
\end{equation}

\noindent where $W_{ij}$ is the weight of the link between node $i$ and $j$, $k_i=\sum_j W_{ij}$ sums the weights of the links associated with node $i$, $c_i$ is the community to which node $i$ is assigned to, $m=\frac{1}{2}\sum_{ij}W_{ij}$, and the $\delta(u, v)=1$ if $u=v$ and 0 otherwise.

To better understand the identification of failure-impact graph using community detection, an illustrating example is depicted in Fig.~\ref{fig:incident_graph_framework} (phase two). In this case, except for nodes $B$ and $F$, other nodes all report incidents. By conducting community detection, we obtain two communities: $\{A, B, C\}$ and $\{C, E, F, G\}$, which are regarded as the complete impact graph of their respective failure. The weight between nodes is provided with their link. %, which is the similarity graph. 
We can see that intra-community links all have a relatively large weight. Such partition can achieve the best modularity score for this example. Particularly, node $H$ is excluded from the second community due to the small weight of its connection to node $F$.

% \hy{why 0.3 is small and should be removed? any threshold }
% \hy{need also mention "complete graph" somewhere here}

To apply community detection, the weight between two nodes should be defined. Inspired by~\cite{liu2019fluxrank}, we combine KPIs with incidents to calculate the behavioral similarity between two nodes and use the similarity value as the weight. Specifically, the weight is composed of two parts, i.e., incident similarity and KPI trend similarity. %\hy{need also say explicitly how "complete" graph is produced}

\subsubsection{Incident similarity}

Incident similarity is to compare the incidents reported by two nodes. Typically, if two nodes encounter similar errors, they will render similar types of incidents. Jaccard index is employed to quantify such similarity, which is defined as the size of the intersection divided by the size of the union of two incident sets:

\begin{equation}
    Jaccard(i, j)=\frac{|inc(i) \cap inc(j)|}{|inc(i) \cup inc(j)|}
\end{equation}

\noindent where $inc(i)$ is the incidents reported by node $i$. In particular, we allow duplicate types of incidents in each set by assigning them a unique number. This is because the distribution of incident types also characterizes the failure symptoms.

\subsubsection{KPI trend similarity} 
\begin{figure}
    \centering
    \includegraphics[width=0.9\linewidth]{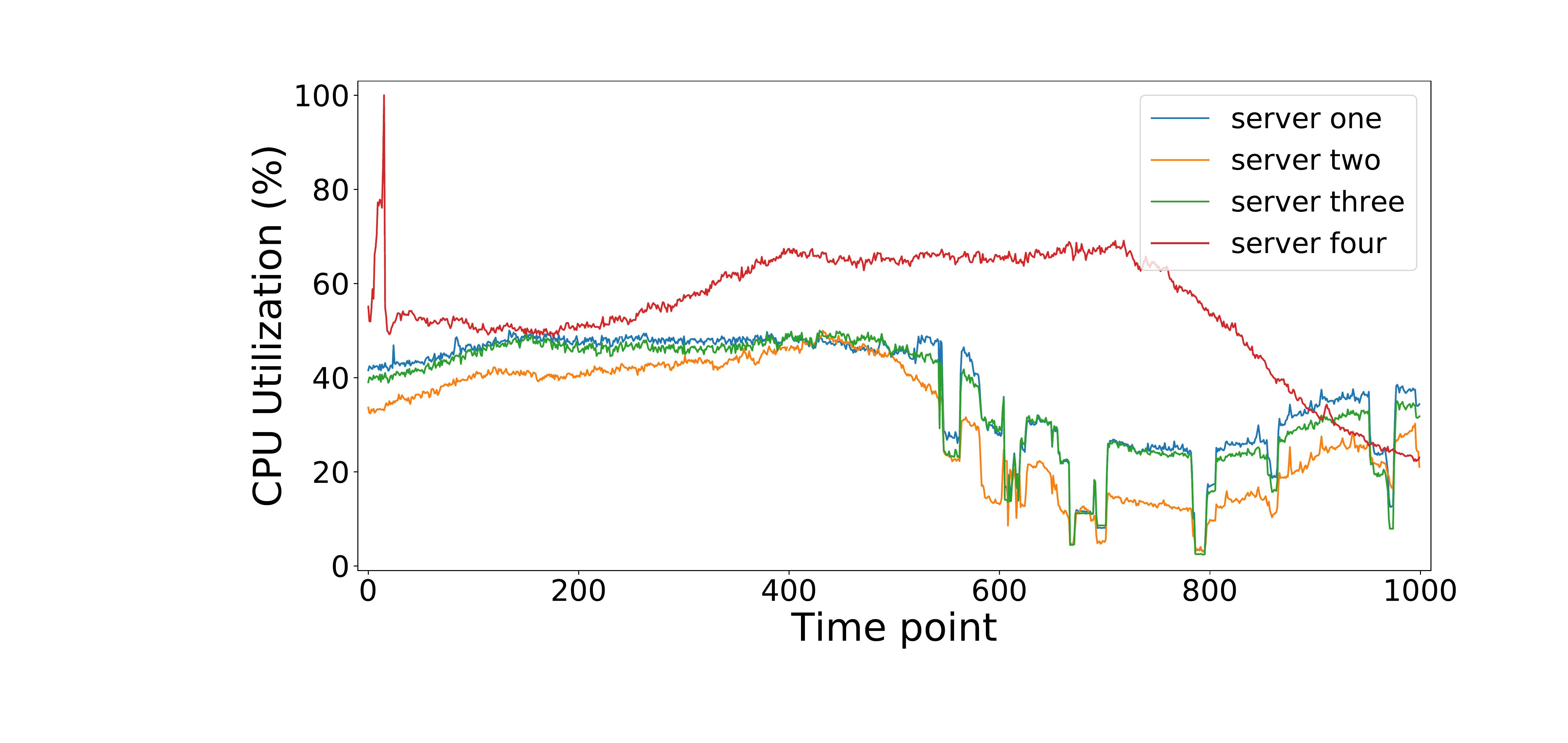}
    \caption{CPU usage curve of four servers}
    \label{fig:kpi_similarity}
\end{figure}

As discussed in Section~\ref{sec:background_problem_statement}, some services may remain silent when failures happen, hindering the tracking of related incidents. To bridge this gap, we resort to KPIs, which are more sophisticated monitoring signals. Intuitively, the KPI trend similarity measures the underlying consistency of cloud components' abnormal behaviors, which cannot be captured by incidents alone. An example is shown in Fig.~\ref{fig:kpi_similarity}, which records the CPU utilization of four servers. Clearly, the curve of the first three servers exhibits a highly similar trend, while such a trend cannot be observed in server four. The implication is that the first three servers are likely to be suffering from the same issue and thus should belong to the same community. We adopt dynamic time warping (DTW)~\cite{keogh2005exact} to measure the similarity between two temporal sequences with varying speeds. We observe the issue of temporal drift between two time series, which is common as different cloud components may not be affected by a failure %receive failure's impact 
simultaneously during its propagation. Therefore, DTW fits our scenario.

The remaining problem is which KPIs should be utilized for similarity evaluation. Normal KPIs which record the system's normal status should be excluded as they provide trivial and noisy information. Therefore, EVT introduced in phase one is utilized again to detect anomalies for each KPI. Only the abnormal KPIs shared by two connected cloud components will be compared. Particularly, when there exists more than one type of abnormal KPIs, we use the average similarity score calculated as follows:

\begin{equation}
    DTW(i, j)=\frac{1}{K}\sum_{k=1}^{K} dtw(t_k^i, t_k^j)
\end{equation}

\noindent where $K$ is the number of KPIs to compare for node $i$ and $j$, $t_k^i$ is the $k^{th}$ KPI of node $i$, and $dtw(u, v)$ measures the DTW similarity between two KPI time series $u$ and $v$, which is normalized for path length. The weight $W_{ij}$ between node $i$ and $j$ is computed by taking the weighted sum of the two types of similarities as follows:

\begin{equation}
    W_{ij}=\alpha \times Jaccard(i, j) + (1-\alpha)\times DTW(i, j)
\end{equation}

\noindent where the balance weight $\alpha$ is a hyper-parameter. In our experiments, if two nodes both report incidents, we set it as 0.5; otherwise, it is set to be 0, i.e., only the KPI trend similarity is considered. 

Finally, for each discovered community, the incidents inside it %are regarded as 
form the complete impact graph of the service failure. %We also set a threshold $10$ for the number of incidents in each graph, to filter out very small communities. This value is set according to the experiences of field engineers.
% \hy{which is $\beta$ ? also say briefly why 50 is set - for example, according to the experiences of domain experts/field engineers, etc}.
Note that in online scenarios, we cannot directly adopt the techniques introduced in this phase for incident aggregation. This is because they involve a comparison between different KPIs, which are not complete until the failures fully manifest themselves. Thus, the comparison is often delayed and inefficient. Moreover, they can be error-prone without fully considering the historical cases.

\subsection{Graph-based Incident Representation Learning}
\label{sec:graph_based_irl}

After obtaining the  %\hy{actually, from the above descriptions, how the "complete" graph is produced needs to made more clearer/explicit} 
impact graph for each service failure (i.e., the actual incidents triggered by it), we can learn the correlations among incidents. Such correlations describe the sets of incidents that tend to appear together.
% Such associated incident sets can help engineers quickly understand the failures.
FP-Growth proposed by Han et al.~\cite{han2000mining} is a standard algorithm to mine such frequent item sets. However, our analysis reveals the following drawbacks it possesses for our problem:

\begin{itemize}
    \item It is vulnerable to background noise. In production environments, some simple incidents are constantly being reported, e.g., ``High CPU utilization rate''. These incidents will appear in many transactions (a collection of items that appear together) for FP-Growth. As a result, unrelated incidents might be put into the same frequent item set due to sharing such incidents. These simple incidents cannot be trivially removed as they provide necessary information about a system, and a burst of such incidents can also indicate serious problems.
    % \hy{not clear: connect incidents belonging to different sets}.
    
    \item It cannot handle incidents with a low frequency. FP-Growth has a parameter called \textit{support}, which describes how frequently an item set is in the dataset. Incident sets with a low support value will be excluded to guarantee the statistical significance of the results. However, more often than not, such incident sets are more important, as they report some critical failures that do not happen frequently.
    
    % \item \textcolor{red}{Even the frequent incident sets can be correctly learned, they do not directly support root cause analysis. This is because in FP-Growth, each incident is only represented with a unique ID, which does not carry specific information of the incident. Thus, it is hard to distinguish important incidents from trivial ones.} \zb{We don't do it, other weak?}
\end{itemize}

In online service systems, different resources (e.g., microservices and devices) are naturally structured in graphical forms, such as service dependency and network IP routing. Therefore, graph representation learning~\cite{hamilton2017representation} can be an ideal solution to deal with the above issues. Graph representation learning is an essential and ubiquitous task with applications ranging from drug design to friendship recommendation in social networks. It aims to find a representation for graph structure that preserves the semantics of the graph. A typical graph representation learning algorithm learns an embedding vector for all nodes of a graph. For example, Chen et al.~\cite{chen2020identifying} employed \textit{node2vec}~\cite{grover2016node2vec} to learn a feature representation for cloud components. Different from them, we propose to learn a representation for each unique type of incident,
% \hy{for each incident or each type of incident? what types?} 
which can appear in multiple places of the graph. In our framework, we employ \textit{DeepWalk}~\cite{perozzi2014deepwalk} because of its simplicity and superior performance. DeepWalk belongs to the class of shallow embedding approaches that learn the node embeddings based on random walk statistics. The basic idea is to learn an embedding $\vartheta_i$ for node $v_i$ in graph $\mathcal{G}$ such that:

\begin{equation}
\label{eq:p_vi_vj}
    EMB(\vartheta_i, \vartheta_j)\triangleq \frac{e^{\vartheta_i \cdot \vartheta_j}}{\sum_{v_k\in \mathcal{V}} e^{\vartheta_i \cdot \vartheta_k}} \approx p_{\mathcal{G},T}(v_j|v_i)
\end{equation}

\noindent where $\mathcal{V}$ is the set of nodes in the graph and $p_{\mathcal{G},T}(v_j|v_i)$ is the probability of visiting $v_j$ within $T$ hops of distance starting at $v_i$. The loss function to maximize such probability is:

\begin{equation}
    \mathcal{L}=\sum_{(v_i,v_j)\in \mathcal{D}} -log(EMB(\vartheta_i, \vartheta_j))
\end{equation}

\noindent where $\mathcal{D}$ is the training data generated by sampling random walks starting from each node. Readers are referred to the original paper~\cite{perozzi2014deepwalk} for more details.

%Typical usages of graph representation learning 

For each failure-impact graph, incident sequences are generated through random walk starting from every node inside. In reality, each node usually generates more than one incident when failures happen. Our tailored random walk strategy therefore contains two hierarchical steps. In the first step, a node is chosen by performing random walks on node level; in the second step, an incident will be randomly selected from those reported by the chosen node. Duplicate types of incidents in a node will be kept because frequency is also an important feature of incidents (it impacts the probability of being selected). Following the original setting of~\cite{grover2016node2vec}, we set the walk length as 40, i.e., each incident sequence will contain 40 samples. Finally, the incident sequences will be fed into a Word2Vec model~\cite{mikolov2013distributed} for embedding vector learning. The Word2Vec model has two important hyper-parameters: the window size and the dimension of the embedding vector. We set the window size as ten by following~\cite{grover2016node2vec} and set the dimension as 128. In particular, by considering the topological distance between incidents, we can alleviate the problem of background noise. This is because as the distance increases, the impact of noisy incidents gradually weakens, while in FP-Growth, all incidents play an equivalent role in a transaction.
% \hy{briefly explain what a transaction is here}. 
% \hy{what is the difference between this and the FSE'20 work [7]? }

\subsection{Online Incident Aggregation}

With the learned incident representation from the last phase, we can conduct incident aggregation in production environments, where the incidents come in a streaming manner. Each group of aggregated incidents represents a specific type of service issue, such as hardware issue, network traffic issue, network interface down, etc. The EVT-based method also plays a role in this phase by continuously monitoring the number of incidents per minute. If it alerts a failure, the online incident aggregation will be triggered.
% Specifically, GRLIA detects service failures by monitoring the number of incidents per minute. If the EVT-based method reports a failure, online incident aggregation will be triggered. 
When two incidents, say $i$ and $j$, appear consecutively, GRLIA measures their similarity. If the similarity score is greater than a predefined threshold, they will be grouped together immediately. In particular, the similarity score consists of two parts, i.e., \textit{historical closeness} (HC) and \textit{topological rescaling} (TR), which are defined as follows:

\begin{equation}
\begin{split}
    HC(i, j) &= \frac{\vartheta_i\cdot \vartheta_j}{\|\vartheta_i\|\times \|\vartheta_j\|} \\
    TR(i, j) &= \frac{1}{max(1, d(i, j) - \mathcal{T})}
\end{split}
\end{equation}

\noindent where $\vartheta_i$ and $\vartheta_j$ are the embedding vectors of incident $i$ and $j$ (as described in Section~\ref{sec:graph_based_irl}),
% \hy{here, are the embeddings for nodes or incidents?}
% zb: for incidents
respectively; $d(i, j)$ is the topological distance between $i$ and $j$, which is the number of hops along their shortest path in the system topology; and $\mathcal{T}$ is the threshold for considering the penalty of long distance.
% \hy{not clear: the threshold to consider the factor of topological distance}
That is, the topological rescaling becomes effective (i.e., $<$1) only if their distance is larger than $\mathcal{T}$. %Setting such a threshold is for the consideration of incomplete impact graph in online scenario.
% \hy{say how it is set. also does it affect your experimental results}.
In our experiments, $\mathcal{T}$ is set as four. Incorrect correlations will be learned if $\mathcal{T}$ is too large, while important correlations will be missed if $\mathcal{T}$ is too small. Our experiments show similar results when $\mathcal{T}$ is in $[3, 6]$. Cosine similarity is adopted to calculate the historical closeness, which is related to their co-occurrences in the past. Finally, the similarity between $i$ and $j$ can be obtained by taking the product of $TR(i, j)$ and $HC(i, j)$:

\begin{equation}
\begin{split}
    sim(i, j)&=TR(i, j)\times HC(i, j) \\
    &=\frac{1}{max(1, d(i, j) - \mathcal{T})}\times \frac{\vartheta_i\cdot \vartheta_j}{\|\vartheta_i\|\times \|\vartheta_j\|}
\end{split}
\end{equation}

We set an aggregation threshold $\lambda$ for $sim(i, j)$ to consider whether or not two incidents are correlated:

\begin{equation}
    cor(i, j)=
    \begin{cases}
    1, & if~sim(i, j) \ge \lambda \\
    0, & otherwise
    \end{cases}
\end{equation}

% \hy{which $\theta$? where is it used? any clustering algorithms used for aggregation? also, say how it is set. also does it affect your experimental results}

\noindent In our experiments,  $\lambda$ is empirically set as 0.7. In particular, the distance of an incident to a group of incidents is defined as the largest value obtained through element-wise comparison.

% \textbf{Root cause recommendation} is to find a small subset of aggregated incidents that best describes the root cause of a failure. An important observation is that critical incidents do not occur frequently and they normally indicate noteworthy failure symptoms. As a result, these incidents would deviate from the majority of incidents in a cloud failure. Thus, for root cause recommendation, we employ the concept of outlier detection to find the incidents of interest. Many algorithms are available for this purpose, such as isolation forests, local outlier factor, one-class SVM, etc. In this report, SVDD (Support Vector Data Description)~\cite{tax2004support} is adopted, which is a one class classification algorithm. SVDD obtains a spherically shaped boundary around a dataset, which separates samples that deviate from the majority of the data by a large margin. In our experiments, incidents predicted as outliers will be regarded as root cause cases.
\section{Experiments}
\label{sec:evaluation}

In this section, we evaluate our framework using real-world incidents collected from industry. Particularly, we aim at answering the following research questions.

\textbf{RQ1}: How effective is the service failure detection module of GRLIA?

\textbf{RQ2}: How effective is GRLIA in incident aggregation? 

\textbf{RQ3}: Can the %completion of
failure-impact graph help incident aggregation?

%\textbf{RQ4}: How effective is GRLIA in assisting incident management?\hy{it is not a RQ, may describe it in experience/success story}

\subsection{Experiment Setting}
\label{sec:exp_setting}

\subsubsection{Dataset}
\label{sec:dataset}

Incident aggregation is a typical problem across different online service systems. In this experiment, we select a representative, large-scale system, i.e., the Networking service of Huawei Cloud, to evaluate the proposed framework. %, which is of a large-scale infrastructure. 
Besides offering traditional services such as Virtual Network, VPN Gateway, it also features intelligent IP networks and other next-generation network solutions.
% As a fundamental service, it supports thousands of other services such as web service and AI platforms. Thus, it has a matured system architecture and reports more abundant and complete incidents. \hy{not easy to explain - if it is matured, why "abundant incidents"...}
In particular, the service system comprises a large and complex topological structure.
% \hy{network topology? topological structure? double check the usage of 'topology'}
In the layer of infrastructure, platform, and software, it has multiple instances of virtual machines, containers, and applications, respectively. In each layer, their dependencies form a topology graph. The cross-layer topology is mainly constructed by their placement relationships, i.e., the mappings between applications, containers, and virtual machines. Like other cloud enterprises, Huawei Cloud's resources are hosted in multiple regions and endpoints worldwide. Each region is composed of several availability zones (isolated locations within regions from which public online services originate and operate) for service reliability assurance. The incident management of the Networking service is also conducted in such a multi-region way, with each region having relatively isolated issues. In this paper, we collect incidents generated between May 2020 and November 2020, during which the Networking service reported a large number of incidents. Although we conduct the evaluation on a single online service system, we believe GRLIA can be easily applied to other online service systems and bring them benefits.

To evaluate the effectiveness of GRLIA, experienced domain engineers %are invited to 
manually labeled related incidents. Thanks to the well-designed incident management system with user-friendly interfaces, the engineers can quickly perform the labeling. Note that the manual labels are only required for evaluating the effectiveness of our framework, which is unsupervised. To calculate the KPI trend similarity, we adopt the following KPIs, which are suggested by the engineers:

\begin{itemize}
    \item \textit{CPU utilization} refers to the amount of processing resources used.
    \item \textit{Round-trip delay} records the amount of time it takes to send a data packet plus the time it takes to receive an acknowledgement of that data packet.
    \item \textit{Port in-bound/out-bound traffic rate} refers to the average amount of data coming-in to/going-out of a port.
    \item \textit{In-bound packet error rate} calculates the error rate of the packet that a network interface receives.
    \item \textit{Out-bound packet loss rate} calculates the loss rate of the packet that a network interface sends.
\end{itemize}

These KPIs are representative that characterize the basic states of the Networking service system. In particular, CPU utilization is monitored for different containers and virtual machines, while the remaining KPIs are monitored for the virtual interfaces of each network device. Each KPI is calculated or sampled every minute. We collect two hours of data to measure the KPI trend similarity. Note that the set of KPIs can be tailored for different systems. For example, a database service may also care about the number of failed database  connection attempts, the number of SQL queries, etc.
% CPU and storage utilization, the number of successful and failed database logon and connection attempts, database operations, SQL queries, and transactions, and so on.

In this paper, we select the largest ten availability zones for experiments, each of which contains a large system topology. Six months of production incidents are collected from the Networking service of Huawei Cloud. The number of distinct incident types is more than 3,000. Similar to~\cite{zhao2020understanding,he2018identifying,lin2018predicting}, we conduct three groups of experiments using incidents reported in the first four months, the first five months, and all months, respectively. In all periods, incident aggregation is applied to the failures that happened in the last month based on the incident representations learned from previous months. Table~\ref{fig:data_statistics} summarizes the dataset. For column \textit{\#Incidents} (resp. \textit{\#Failures}), the first figure calculates the incidents (resp. failures) captured during the training period, while the second figure shows that of the evaluation month. Particularly, some failures are of small scale and can be quickly mitigated, while others are cross-region and become an expensive drain on the company's revenue. We can see each failure is associated with roughly 200 incidents, demonstrating a strong need for incident aggregation.

% \hy{also, say the values of various thresholds used in the experiments, and other parameter settings.}

\begin{table}
    \centering
    \caption{Dataset statistics}
    \includegraphics[width=1.0\linewidth]{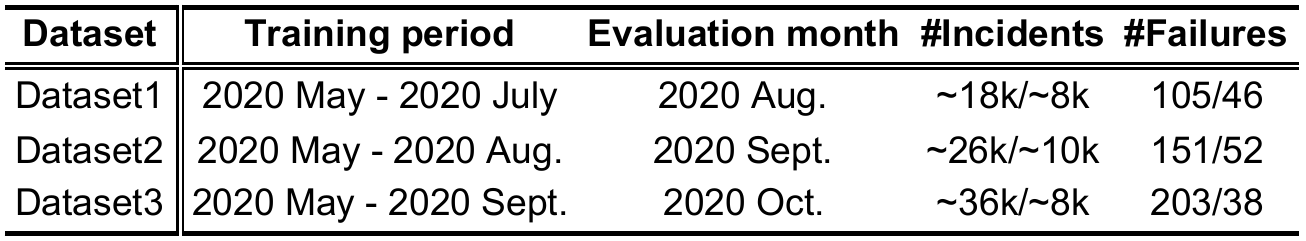}
    \label{fig:data_statistics}
    \vspace{-10pt}
\end{table}

% \begin{table}
%     \centering
%     \caption{Dataset separation}
%     \includegraphics[width=1.0\linewidth]{images/data_separation.pdf}
%     \label{fig:data_separation}
% \end{table}

\subsubsection{Evaluation Metrics}

For RQ1, which is a binary classification problem, we employ \textit{precision}, \textit{recall}, and \textit{F1 score} for evaluation. Specifically, precision measures the percentage of incident bursts that are successfully identified as service failures over all the incident bursts that are predicted as failures: $precision=\frac{TP}{TP+FP}$. Recall calculates the portion of service failures that are successfully identified by GRLIA over all the actual service failures: $recall=\frac{TP}{TP+FN}$. Finally, F1 score is the harmonic mean of precision and recall: $F1~score=\frac{2\times precision\times recall}{precision+recall}$. $TP$ is the number of service failures that are correctly discovered by GRLIA; $FP$ is the number of trivial incident bursts (i.e., no failure is actually happening)
% \hy{what are benign incidents? can incidents be benign?} 
that are wrongly predicted as service failures by GRLIA; $FN$ is the number of service failures that GRLIA fails to discover.

% $TN$\zb{no used?} is the number of trivial incident bursts that are not identified as service failures by GRLIA.

For RQ2 and RQ3, we choose Normalized Mutual Information (NMI)~\cite{stanford_nmi}, which is a widely used metric for evaluating the quality of clustering algorithms. The value of NMI ranges from 0 to 1 with 0 indicating the worst result (no mutual information) and 1 the best (perfect correlation): $NMI(\Omega, \mathbb{C})=\frac{2\times I(\Omega; \mathbb{C})}{H(\Omega) + H(\mathbb{C})}$, where $\Omega$ is the set of clusters, $\mathbb{C}$ is the set of classes, $H(\cdot)$ is the entropy, and $I(\Omega; \mathbb{C})$ calculates and mutual information between $\Omega$ and $\mathbb{C}$.

% For root cause recommendation, since for each set of aggregated incidents, a small subset of members will be selected as the culprit, it can be regarded as a classification problem. Therefore, Precision, Recall, and F1 score are employed for evaluation. For each set of incidents, Precision measures the percentage of root-cause incidents that are successfully identified over all incidents, Recall calculates the portion of root-cause incidents that are successfully identified over all the actual root-cause incidents, and F1 score is the harmonic mean of Precision and Recall. The final results are reported by averaging over the scores obtained in all set of aggregated incidents.

% \subsubsection{Parameter Settings}

% \textcolor{red}{Any for EVT?}

% \textcolor{red}{DeepWalk: length of the sequence}

% \textcolor{red}{Some hyper-parameters: $\mathbb{T}$}

% \zb{Draw a table? Not quite necessary}

\subsubsection{Implementation}

Our framework is implemented in Python. We parallelize our experiments by assigning availability zones to different processors. The output of each processor is a list of incident sequences generated through random walk, which we merge and feed to a Word2Vec model implemented with Gensim~\cite{gensim}, an open-source library for topic modeling and natural language processing. We run our experiments on a machine with 20 Intel(R) Xeon(R) Gold 6148 CPU @ 2.60GHz, and 256GB of RAM. The results show that each phase of our framework takes only a few seconds.
% \hy{if possible, can give a time range}
The last phase can even produce results in a real-time manner as it only involves simple vector calculation. Thus, our framework can quickly respond in online scenarios. This demonstrates that GRLIA is of high efficiency.

%\subsection{Baselines}
\subsection{Comparative Methods}

The following %baselines
methods are selected for comparative evaluation of GRLIA.

\begin{itemize}
    \item \textit{FP-Growth}~\cite{han2000mining}. FP-Growth is a widely-used algorithm for association pattern mining. It is utilized as an analytical process that finds a set of items that frequently co-occur in datasets. In our experiments, each impact graph is regarded as a transaction for this algorithm. Given a set of impact graphs, it searches incidents that often appear together, regardless of their distance.
    % \item \textit{TF-IDF}. TF-IDF (term frequency–inverse document frequency) is a numerical statistic that measures how important a word is to a document in a collection or corpus. Particularly, it is composed of two parts: term frequency (TF) and inverse document frequency (IDF). The TF counts the occurrence of a word in a document $n_w$, while the IDF evaluates how much information the word provides: $log(\frac{N}{n_w})$ where $N$ is the total number of documents. In this report, it is used to find the most informative incidents in a cascading topology, which are regarded as the root cause. In particular, the number of selected incidents is the same as our framework. Similarly, this method cannot perform incident aggregation.
    \item \textit{UHAS}~\cite{zhao2020understanding}. This approach is proposed by Zhao et al. aiming at handling alert storms for online service systems. Similar to incident bursts, alert storms also serve as a signal for service failures. Particularly, UHAS employs DBSCAN for alert clustering based on their textual and topological similarity. The textual similarity between two alerts is measured by Jaccard distance. The topological similarity considers two types of topologies, i.e., software topology (service) and hardware topology (server). The topological distance is computed by the shortest path length between two nodes. Finally, a weighted combination of the two types of similarities yields the final similarity score.
    
    \item \textit{LiDAR~\cite{chen2020identifying}}. LiDAR is a supervised method proposed by Chen et al. to identify linked incidents in large-scale online service systems. Specifically, LiDAR is composed of two modules, i.e., textual encoding module and component embedding module. The first module produces similar representations for the text description of linked incidents, which are labeled by engineers. In the evaluation stage, the textual similarity between two incidents is measured by the cosine distance of their representations. The second module learns a representation for the system topology (instead of incidents). The final similarity is calculated by taking a weighted sum of both parts. As LiDAR is supervised, it would be unfair to compare it with other unsupervised methods. Considering the success of Word2Vec model~\cite{mikolov2013distributed,mikolov2017advances} in identifying semantically similar words (in an unsupervised manner), we alter LiDAR to be unsupervised to fit our scenario by representing the text of incidents with off-the-shelf word vectors~\cite{joulin2016fasttext}.
\end{itemize}

\subsection{Experimental Results}

\subsubsection{\textbf{RQ1}: The Effectiveness of GRLIA's Service Failure Detection}

To answer this research question, we compare GRLIA with the fixed thresholding method on three datasets and report precision, recall, and F1 score. Thresholding remains an effective way for anomaly detection in production systems and serves as a baseline in many existing work. Since both methods require no parameter training, we use them to detect failures for both the training data and evaluation data. Particularly, the threshold of the baseline method is \#incidents/min$>$50, which is recommended by field engineers. Moreover, the ground truth is obtained directly from the historical failure tickets, which are stored in the incident management system.

The results are shown in Table~\ref{tab:evt}, where GRLIA outperforms the fixed thresholding in all datasets and metrics. In particular, GRLIA achieves an F1 score of more than 0.93 in different datasets, demonstrating its effectiveness in service failure detection. Indeed, we observe that some failures may not always incur a large number of incidents at the beginning. However, if ignored, they could become worse and end up yielding more severe impacts across multiple services. Fixed thresholding does not possess the merit of threshold adaptation based on the context and thus produces many false positives. GRLIA outperforms it for being able to adjust the threshold automatically.
%Particularly, compared to recall, the precision is not very important.
% \hy{could be removed: Particularly, compared to recall, the precision is not very interesting}
% zb: here should be my misleading expression. I intend to say that the precision is not very important
% \hy{"not very interesting" is too vague. Also, your results show that precision values are all better than thresholding. Precision values are much better than Recall values. so perhaps no need to say "not interesting" here}.
% This is because GRLIA is able to learn the correlations
% % \hy{does it learn association rules?}
% between incidents. %The relationship between two incidents will remain valid even they appear together not within a failure. %However, we cannot use arbitrary intervals of incidents, which could induce noise \hy{what does the last sentence mean?}.
% \hy{is this necessary}
% \hy{also, in Table III, are Precision/Recall values correct? Precision values are much better than Recall values}
% zb: yes, since we expect to have a better recall, we sacrifice some precision

\begin{table}
    \centering
    \caption{Experimental results of service failure detection}
    \includegraphics[width=0.84\linewidth]{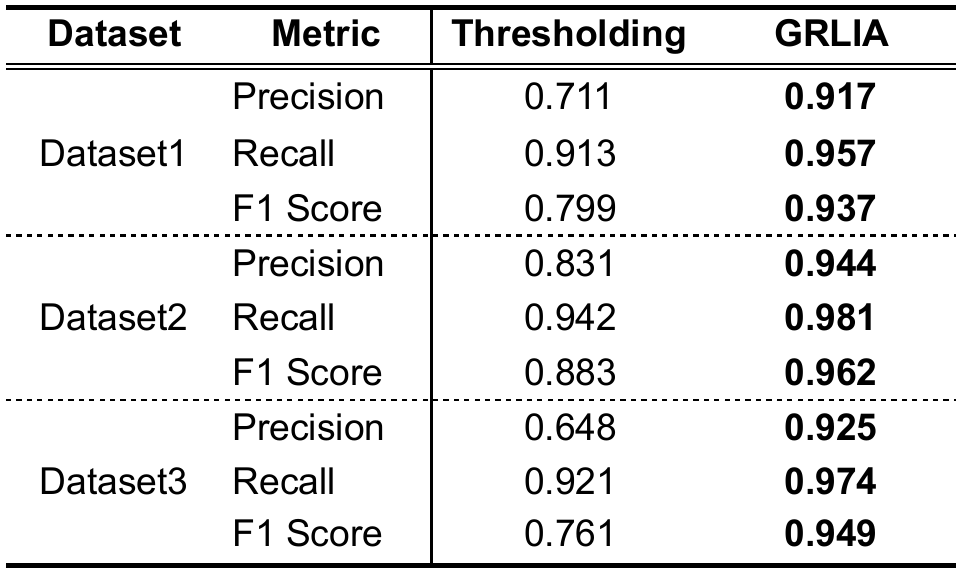}
    \label{tab:evt}
\end{table}

\subsubsection{\textbf{RQ2}: The Effectiveness of GRLIA in Incident Aggregation}

We compare the performance of GRLIA against a series of baseline methods for incident aggregation. Table~\ref{tab:aggregation} shows the NMI values of different experiments. From dataset 1 to 3, GRLIA achieves an NMI score of 0.831, 0.866, and 0.912, respectively, while the best results from the baseline methods are 0.742, 0.758, and 0.826, all attained by LiDAR. LiDAR outperforms UHAS by explicitly considering the entire system topology.
% This demonstrates that GRLIA outperforms all baseline methods by a noticeable margin.
Except for UHAS, all approaches achieve better performance with more training data available. This is because UHAS directly works on alert storms when failures are detected. Without learning from the history, it cannot handle complicated scenarios. Recall that both UHAS and LiDAR rely on the textual similarity between incidents. However, in our system, related incidents do not necessarily possess similar text descriptions. For example, there is a clear correlation between the incident ``Traffic drops sharply in vRouter'' and ``OS network ping abnormal'' in VPC service, which tends to be missed by them. Moreover, monitors that render incidents are configured by multiple service teams, which further damages the credit of textual similarity. This is particularly true for some critical incidents because they are often tailored for special system errors, which may not be shared across different services. On the other hand, although GRLIA does not explicitly leverage incident's textual features, our experiments show that it is capable of correlating incidents that share some common words, e.g., ``VPC service tomcat port does not exist'' and ``VPC service tomcat status is dead''. This is because such a relationship is reflected in their temporal and topological locality, which can be precisely captured by incidents' representation vectors.

Another observation is that FP-Growth does not fit the task of incident aggregation, whose best NMI score is 0.546. As discussed in Section~\ref{sec:graph_based_irl}, this method is not robust against background noise. Indeed, in the system, some trivial incidents (e.g., ``Virtual machine is in abnormal state'') are continuously being reported, which may connect incidents from distinct groups. Furthermore, many essential incidents are excluded by this method due to low frequency, which is undesirable. This problem can be effectively alleviated by leveraging the topological relationship between incidents as done by other approaches. According to Eq.~\ref{eq:p_vi_vj}, the impact of background noise weakens with distance.
% \hy{Equation 5 is about embedding?} \zb{Yes, here I mean if the distance is larger, the probability of $p(v_i|v_j)$ will be smaller, so the impact is weaker}.
However, in FP-Growth, each incident co-occurrence will be counted equally towards the final association rules. UHAS considers the topological similarity by simply calculating the distance. LiDAR employs a more expressive machine learning model, i.e., \textit{node2vec}~\cite{grover2016node2vec}, an algorithmic framework for learning a continuous representation for a network's nodes. However, they both ignore the problem of incomplete failure-impact graph, which is a common issue in online service systems according to our study. The necessity of completing the impact graph will be demonstrated in RQ3. Moreover, different from the traditional applications of graph representation learning, we learn a representation for each unique type of incident, which compactly encodes its relationship with others.

\begin{table}
    \centering
    \caption{Experimental results of incident aggregation}
    \includegraphics[width=0.9\linewidth]{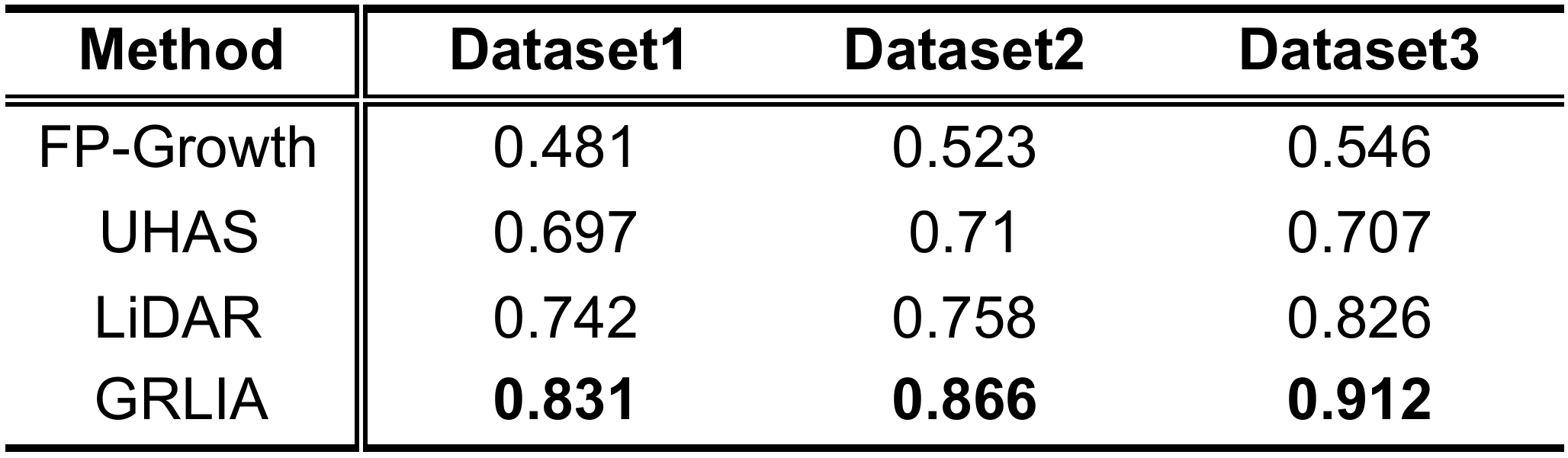}
    \label{tab:aggregation}
\end{table}

\subsubsection{\textbf{RQ3}: The Necessity of the Failure-Impact Graph for Incident Aggregation}

We demonstrate the importance of impact graphs by creating a variant of GRLIA without the phase of failure-impact graph completion (i.e., phase two in Fig.~\ref{fig:incident_graph_framework}), denoted as GRLIA$'$. We follow LiDAR to remove this feature, which considers two incidents as related only when they are directly connected in the system topology. The experimental results are presented in Table~\ref{tab:aggregation_no_impact_graph}, where we can see a noticeable drop in the NMI score for all datasets. Due to the high complexity and large scale of online service systems, monitors are often configured in an ad-hoc manner. These monitors may not be able to accommodate to the ever-changing systems and environments. Thus, some incidents are not successfully captured by them. System engineers may incorrectly perceive the service as healthy, which is a typical situation of gray failures~\cite{huang2017gray}. Without completing the impact graph of failures, the true correlations among incidents cannot be fully recovered.

\begin{table}
    \centering
    \caption{Experimental results of incident aggregation using GRLIA (w/ and w/o failure-impact graph completion)}
    \includegraphics[width=0.9\linewidth]{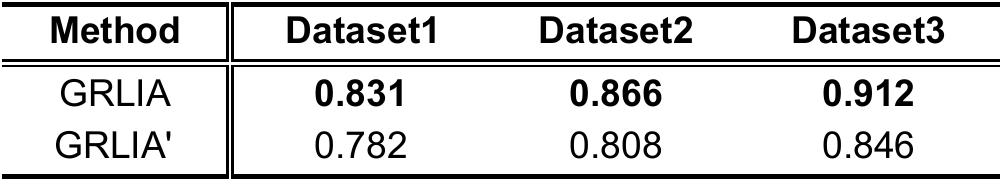}
    \label{tab:aggregation_no_impact_graph}
    \vspace{-6pt}
\end{table}

% \subsubsection{\textbf{RQ4} Effectiveness of GRLIA in assisting incident management}

% GRLIA has been successfully integrated into company $\mathcal{Q}$'s incident management system. To quantify the practical benefits conveyed to the Networking service of company $\mathcal{Q}$, we further collect failure tickets generated during \textcolor{red}{November} 2020 in the studied availability zones. In total, 26 failures are recorded.
% % Each ticket records the detailed information of the failure, e.g., service, timestamp, title, and the complete pipeline of failure handling, e.g., fault localization process, solution.
% We calculate the average failure handling time of November and compare it with that of August, September, and October. Results show the time reduction rate of the comparisons are 24.8\%, 21.9\%, and 18.6\%, respectively, which demonstrate the effectiveness of GRLIA in accelerating incident management of company $\mathcal{Q}$. \zb{Real-world case?}

\subsection{Threats to Validity}

During our study, we have identified the following major threats to the validity.

\textbf{Labeling noise}. Our experiments are conducted based on six months of real-world incidents collected from Huawei Cloud. The evaluation requires engineers to inspect and label the incidents manually. Label noises (false positives/false negatives) may be introduced during the manual labeling process. However, the engineers we invite are cloud system professionals and have years of system troubleshooting experience. Moreover, the labeling work can be done quickly and confidently thanks to the incident management system which has user-friendly interfaces. Therefore, we believe the amount of noise is small (if it exists).

\textbf{Selection of study subjects}. In our experiments, we only collect incidents from one online service of Huawei Cloud, i.e., the Networking service. This is a large-scale service that supports many upper-layer services such as web application, virtual machine. Sufficient data can be collected from this service system. Another benefit we can enjoy is that the topology of the Networking service system is readily available and accurate. Although we use the Networking service as the subject in this paper, our proposed framework is generalizable, as this service is a typical, representative online service. Thus, we believe GRLIA can be applied to other services and cloud computing platforms and bring them benefits.
%While previous studies \hy{cite} includes a step of service dependency discovery, which may include noise.

The second type of subject that could threaten the validity is the KPI. In production systems, there is a large amount of KPIs available to gauge the similarity between two nodes. Although we only select six representative KPIs (as presented in Section~\ref{sec:dataset}), they record the basic and critical states of a service component. Thus, we believe they are able to profile the service system comprehensively.

%\hy{can also say these KPIs are common to all cloud service systems, therefore the proposed framework is generalizable. also say the Networking service we studied is also a representative cloud service. Although the experiment is on the Networking service, the results are generalizable and GRLIA can be applied to other service systems...}

\textbf{Implementation and parameter setting}. The implementation and parameter setting are two critical internal threats to the validity. To reduce the threat of implementation, we employ peer code review. Specifically, the authors are invited to carefully check others' code for mistakes. In terms of parameter setting, we conduct many groups of comparative experiments with different parameters. We choose the parameters by following the original work or empirically based on the best experimental results. In particular, we found GRLIA is not very sensitive to the parameter setting.

\section{Discussion}
\label{sec:discussion}

\subsection{Success Story}

GRLIA has been successfully incorporated into the incident management system of Huawei Cloud. %We have seen it shedding lights on the pipeline of incident handling. 
Based on the positive feedback we have received, on-site engineers (OSEs) highly appreciated the novelty of our approach and benefited from it during their daily system maintenance.
Specifically, OSEs confirmed the difficulty of the auto-detection of service failures in the existing monitoring system. This is because simple detection techniques (e.g., fixed thresholding) are widely adopted.
%, TF-IDF weighting are dominating the system. 
GRLIA introduces more intelligence and automation by leveraging EVT-based incident burst detection. Interestingly, OSEs found problems for some monitors by comparing their configurations with the aggregated incidents, including wrong names, missing information, etc. Meanwhile, during failure diagnosis, incident aggregation assists OSEs in reducing their investigation scope. Before the deployment of GRLIA, they would have to examine a large number of incidents to locate the failures.

To quantify the practical benefits conveyed to the Networking service system, we further collect failure tickets generated during November 2020.  %in the studied availability zones. 
 In total, 26 failures are recorded.
% Each ticket records the detailed information of the failure, e.g., service, timestamp, title, and the complete pipeline of failure handling, e.g., fault localization process, solution.
We calculate the average failure handling time in November and compare it with that in August, September, and October. Results show that the time reduction rate is 24.8\%, 21.9\%, and 18.6\%, respectively, demonstrating the effectiveness of GRLIA in accelerating the incident management of Huawei Cloud.

\subsection{Lessons Learned}

\textbf{Optimizing monitor configurations}. Today, popular online services are %have reached the scale of 
serving tens of millions of customers. During daily operations, they can produce terabytes and even petabytes of telemetry data such as KPIs, logs, and incidents. However, the majority of these data does not contain much valuable information for service failure analysis. For example, a significant portion of KPIs only record plain system runtime states; most of the incidents are trivial and likely to mitigate automatically with time. The configuration of system monitors should be optimized to report more important yet fewer incidents. In the meantime, monitor configurations show different styles across different service teams, making the monitoring data heterogeneous.
% \hy{why mentioning feature extraction from raw incidents here}.
Standards should be established for monitor configurations so that high-quality incidents can be created to facilitate the follow-up system analysis, e.g., fault localization.

\textbf{Building data collection pipeline}. In online service systems, IT operations play a critical role in system maintenance. Since it is data-driven by nature, modern cloud service providers should build a complete and efficient pipeline for monitoring data collection. Common data quality issues include extremely imbalanced data, small quantity of data, poor signal-to-noise ratio, etc. In general, we are facing the following three challenges: \textit{1) What data should be collected?} We need to identify what metrics and events that are most representative for cloud resource health. Not everything that can be measured needs to be monitored. \textit{2) How to collect and label data?} Labeling incidents (e.g., incident linkages, culprit incidents) requires OSEs to have a decent knowledge about the cloud systems. Since they often devote themselves to emerging issue mitigation and resolution, tools should be developed to facilitate the labeling process, such as label recommendation and friendly interfaces. \textit{3) How to store and query data?} Today's cloud monitoring data are challenging the conventional database systems. To save space, domain-specific compression techniques should be developed, for example, log compression~\cite{liu2019logzip,christensen2013adaptive,he2018characterizing}.

\section{Related Work}
\label{sec:related_work}

%In the literature, there are tremendous efforts devoted to improve the quality and reliability of online service systems. These works exploit different types of system monitoring data, e.g., KPIs, logs, alerts, incidents. Based on the tasks, we classify them into the following two categories.

\subsection{Problem Identification}
% Problem Identification
To provide high-quality online services, many researchers have conducted a series of investigations, including problem identification and incident diagnosis from runtime log data and alerts~\cite{lin2016log, he2018identifying, jiang2011ranking}.
% log
For example, to identify problems from a large volume of log data, Lin et al.~\cite{lin2016log} proposed LogCluster to cluster log sequences and pick the center of each cluster.
Rosenberg et al.~\cite{rosenberg2018improving} extended LogCluster by incorporating dimension reduction techniques to solve the high-dimension challenge of log sequence vectors.
% alert
Inspired by LogCluster~\cite{lin2016log}, Zhao et al.~\cite{zhao2020understanding} clustered online service alerts to identify the representative alerts to engineers.
Different from the clustering techniques, Jiang et al.~\cite{jiang2011ranking} proposed an alert prioritization approach by ranking the importance of alerts based on the KPIs in alert data.
The top-ranked alerts are more valuable to identifying problems.
However, this approach has a limited scope of application because it is only practical to KPI alerts generated from manually defined threshold rules.
% incident diagnosis
To conduct problem identification more aggressively, Chen et al.~\cite{chen2019outage} proposed an incident diagnosis framework to predict general incidents by analyzing their relationships with different alerting signals.
Zhao et al.~\cite{zhao2020real} considered a more practical scenario where there are plenty of noisy alerts in online service systems.
They proposed eWarn to filter out the noisy alerts and generate interpretable results for incident prediction.

\subsection{Incident Management}
% Incident Management
In recent years, cloud computing has gained unprecedented popularity, and incidents are almost inevitable.
Thus, incident management becomes a hotspot topic in both academia and industry.
Massive amount of effort has been devoted to incident detection~\cite{lin2018predicting,zhao2020real,gu2020efficient,lin2014unveiling} and incident triage~\cite{gu2020efficient,gao2020scouts,chen2019empirical,chen2019continuous}.
 % incident detection
%Kikuchi et al.~\cite{kikuchi2015prediction} proposed a text mining method to predict the workload of resolving incidents in incident management.
For example, Lim et al.~\cite{lim2014identifying} utilized Hidden Markov Random Field for performance issue clustering to identify representative issues. % to speed up the troubleshooting process.
% reduce number of incidents
Chen et al.~\cite{chen2019continuous} proposed DeepCT, a deep learning-based approach that is able to accumulate knowledge from incidents' discussions and automate incident triage. 
However, due to high manual examination costs, 
%due to high complexity and manual examine for root cause, 
these methods cannot handle the overwhelming number of incidents.
Many existing work~\cite{xu2017lightweight, zhao2020understanding} address this problem by reducing the duplicated or correlated alerts.
For example, Zhao et al.~\cite{zhao2020automatically} aimed to recommend the severe alerts to engineers.
Lin et al.~\cite{lin2014unveiling} proposed an alert correlation method to cluster semi-structured alert texts to gain insights from the clustering results.

 % textual similairty
Similar to our method, Zhao et al.~\cite{zhao2020understanding} conducted alert reduction by calculating their textual and topological similarity. The centroid alert of each cluster is then selected as the representative incident to engineers.
Specifically, they first leveraged conventional methods to detect alert storms and the associated anomalous alerts, and then adopted DBSCAN~\cite{ester1996density} to cluster alerts based on their textual and topological similarity.
% semantic embedding
Another similar work is LiDAR proposed by Chen et al.~\cite{chen2020identifying}, which links relevant incidents by incorporating the representation of cloud components.
Their framework consists of two modules, a textual encoding module and a component embedding module.
The first module learns a representation vector for incident’s description in a supervised manner. The textual similarity between two incidents is measured by the cosine distance of their representation vectors.
Similarly, the second module learns a vector for system components.
The final similarity is calculated by leveraging two parts of information.
However, these methods employ a simple weighted sum to combine the information from different sources, and still hardly capture the relationship between incidents.
Differently, our method utilizes sophisticated graph representation learning to obtain the semantic relationship of incidents from diverse sources, including temporal locality, topological structure, and KPI metric data.
% un-supervised
Moreover, many existing incident management methods rely on supervised machine learning techniques to detect anomalies or conduct incident triage.
More intelligent approaches with weak-supervision or even unsupervised frameworks are still largely unexplored.

\section{Conclusion}
\label{sec:conclusion}

%\zb{To be done by Zhuangbin}

In this paper, we propose GRLIA, an incident aggregation framework based on graph representation learning. The representation for different types of incidents is learned in an unsupervised and unified fashion, which encodes the interactions among incidents in both temporal and topological dimensions. Online incident aggregation can be efficiently performed by calculating their distance. We have conducted experiments with real-world incidents collected from Huawei Cloud. Compared with fixed thresholding, GRLIA achieves better performance for being able to adjust the threshold automatically. In terms of online incident aggregation, GRLIA also outperforms existing methods by a noticeable margin, confirming its effectiveness. Furthermore, our framework has been successfully incorporated into the incident management system of Huawei Cloud. Feedback from on-site engineers confirms its practical usefulness. We believe our proposed incident aggregation framework can assist engineers in failure understanding and diagnosis.

% \textcolor{blue}{Our source code and sample experimental data are available on GitHub: {\url{https://github.com/CodeReviewAnonymous/ASE21-Submission166}}.}
% \hy{mark the changes in BLUE, as required in their email. Also, only one Python source file in the webpage? also, better to include a Readme, etc} \zb{Ok, will do it}

% \hy{release the code and sample data? like in the DSN submission?}

% \hy{the reference [9] has a wrong year} \zb{I double checked and it is 2017?}\hy{sorry, it was reference [5], now it seems fixed} \zb{Ok, thanks}

% \hy{not sure if all major comments are addressed. also, need a reply letter.}
%\hy{better do it now. also, ASE allows supplementary materials}

%For future work, we will extend our framework to automate root cause localization by leveraging the learned correlations among incidents. We will also investigate the causal relationships among the incidents.

%To this end, we need to have a clear view of the causal relationship between incidents, which requires more 

% {Our source code and sample experimental data are publicly available on GitHub: \url{https://github.com/dsn21-85/code}.}

% \hy{I think ASE requirs 10+2 pages. So need to cut some. Also, the Repo name could be updated}

% \hy{the DSN CFP says: we expect all papers to provide enough detail to enable reproducibility of their experimental results and encourage authors, whenever possible, to make both the software developed in the experiments or datasets publicly available.}\yx{add code link in implementation part}\hy{if possible, can provide some sample data (no need to be complete), which can work with the tool}
\section*{Acknowledgement}

The work was supported by Key-Area Research and Development Program of Guangdong Province (No. 2020B010165002), the Research Grants Council of the Hong Kong SAR, China (CUHK 14210920), and Australian Research Council (ARC) Discovery Project (DP200102940). %\hy{Michael and I have this joint ARC project.}

\balance
\bibliographystyle{IEEEtran}
% \balance
\bibliography{bibliography}
\balance
\end{document}